\documentclass{ieeeaccess}
\usepackage{cite}
\usepackage{amsmath,amssymb,amsfonts}
\usepackage{algorithmic}
\usepackage{graphicx}
\usepackage{textcomp}
\def\BibTeX{{\rm B\kern-.05em{\sc i\kern-.025em b}\kern-.08em
    T\kern-.1667em\lower.7ex\hbox{E}\kern-.125emX}}

\usepackage[font={sf,scriptsize},
            labelfont={bf},
            justification= raggedright, 
            singlelinecheck=false]
            {caption}
            
\usepackage[caption=false,font=footnotesize]{subfig}

\begin{document}
\history{Date of publication xxxx 00, 0000, date of current version xxxx 00, 0000.}
\doi{10.1109/ACCESS.2022.DOI}

\title{Bayesian Neural Networks for\\Reversible Steganography}
\author{\uppercase{Ching-Chun Chang}\authorrefmark{}}
\address[]{Department of Computer Science, University of Warwick, Coventry, UK} 

\markboth
{C.-C. Chang: Bayesian Neural Networks for Reversible Steganography}
{C.-C. Chang: Bayesian Neural Networks for Reversible Steganography}

\corresp{Corresponding author: C.-C. Chang (e-mail: c.c.chang@warwickgrad.net).}

\begin{abstract}
Recent advances in deep learning have led to a paradigm shift in the field of reversible steganography. A fundamental pillar of reversible steganography is predictive modelling which can be realised via deep neural networks. However, non-trivial errors exist in inferences about some out-of-distribution and noisy data. In view of this issue, we propose to consider uncertainty in predictive models based upon a theoretical framework of Bayesian deep learning, thereby creating an adaptive steganographic system. Most modern deep-learning models are regarded as deterministic because they only offer predictions while failing to provide uncertainty measurement. Bayesian neural networks bring a probabilistic perspective to deep learning and can be regarded as self-aware intelligent machinery; that is, a machine that knows its own limitations. To quantify uncertainty, we apply Bayesian statistics to model the predictive distribution and approximate it through Monte Carlo sampling with stochastic forward passes. We further show that predictive uncertainty can be disentangled into aleatoric and epistemic uncertainties and these quantities can be learnt unsupervised. Experimental results demonstrate an improvement delivered by Bayesian uncertainty analysis upon steganographic rate-distortion performance.
\end{abstract}


\titlepgskip=-15pt

\maketitle

\section{Introduction}
\label{sec:introduction}
\PARstart{A}{rtificial} intelligence arises from the question `Can machines think?'~\cite{10.1093/mind/LIX.236.433}. Machine learning refocuses attention on addressing solvable problems of a practical nature automatically through the use of data. Deep learning is a subclass of machine-learning algorithms based on neural networks and connectionism~\cite{NIPS2012_AlexNet, Simonyan15, LeCun:2015aa}. The advent of data-centric artificial intelligence is accompanied by cybersecurity concerns. It has been reported that machine-learning models are vulnerable to adversarial attacks~\cite{10.1145/3134599}. One such example is the introduction of invisible perturbations to data to cause erroneous decision-making~\cite{2015_Perturb_Goodfellow}. Another example is the injection of poisonous data that subsequently contaminates and disrupts the learning process~\cite{Poisoning17}. There is also an insidious threat that malware codes could be hidden within a neural network model using obfuscation techniques and subsequently triggered by specific input data~\cite{10.1145/3427228.3427268} (metaphorically a Trojan Horse stratagem). Data authentication serves a crucial role in cybersecurity, being at the forefront of defensive strategies against various cybercrimes. Modern cryptography offers a variety of approaches to authentication (e.g. digital signatures~\cite{10.1145/359340.359342} and trusted timestamps~\cite{Haber:1991aa}). However, the maintenance of such auxiliary information imposes an additional burden upon data management requirements.

Steganography is the art and science of covering messages within digital media~\cite{668971}. It can serve as a potential solution for managing auxiliary information. By embedding an auxiliary message into its corresponding data sample, the link between the sample and the message is naturally preserved throughout the entire information lifecycle. While steganographic distortion is generally imperceptible, \emph{reversibility} is required for applications in which data integrity is a major priority (e.g. forensic science, legal proceedings, medical diagnosis, and military reconnaissance)~\cite{2001_Fridrich_Invertible, 2003_1227616, 2003_1196739, 2006_1608163, 2007_4291553, 2016_RDH_Survey}. When a steganographic process is irreversible, distortion may accumulate over time and eventually render data worthless. The recent development of deep learning has brought a paradigm shift in reversible steganography~\cite{Duan:2019aa, Jung:2019aa, Lu_2021_CVPR}. Similar to lossless compression~\cite{1948_6773024}, predictive modelling forms a fundamental pillar of reversible steganography for analysing redundancy in digital signals. It has been reported that deep neural networks can be used as advanced predictive models~\cite{2020_9245471, Hu:2021aa, Chang:2021aa}. Despite the improved accuracy offered by neural networks, non-trivial prediction errors still occur when making inferences about some \emph{out-of-distribution} and \emph{noisy} test data. This leads us to the study of uncertainty in deep learning.

Most modern deep-learning models are regarded as \emph{deterministic} as they only offer predictions while lacking confidence bounds for data analysis and decision-making, thereby incurring risks in automated systems. As a safety concern, it is important to be aware of the limitations of a machine that is deployed in real-world settings and granted autonomous control~\cite{ijcai2017-661}. While the notion of machine consciousness is illusive and there is no indication that contemporary artificial intelligence is anywhere close to engendering that, uncertainty quantification would be a principal requirement for the development of self-aware intelligent machinery. Bayesian deep learning provides a way of calculating uncertainty based on a \emph{probabilistic} conception~\cite{10.1145/3409383}. 

In this paper, we study predictive uncertainty in deep neural networks for reversible steganography based on a theoretical framework of Bayesian deep learning~\cite{Gal2016Uncertainty}. Our objective is to develop a learning-based method for quantifying uncertainty caused by out-of-distribution and noisy data, thereby enabling an adaptive steganographic system and improving steganographic rate-distortion performance.




\section{Methodology}
We begin by formulating a reversible steganographic scheme that incorporates a Bayesian neural network and then present a derivation of uncertainty.

\subsection{Reversible Steganography}
\begin{figure}[t]
\centerline{\includegraphics[width=0.85\columnwidth]{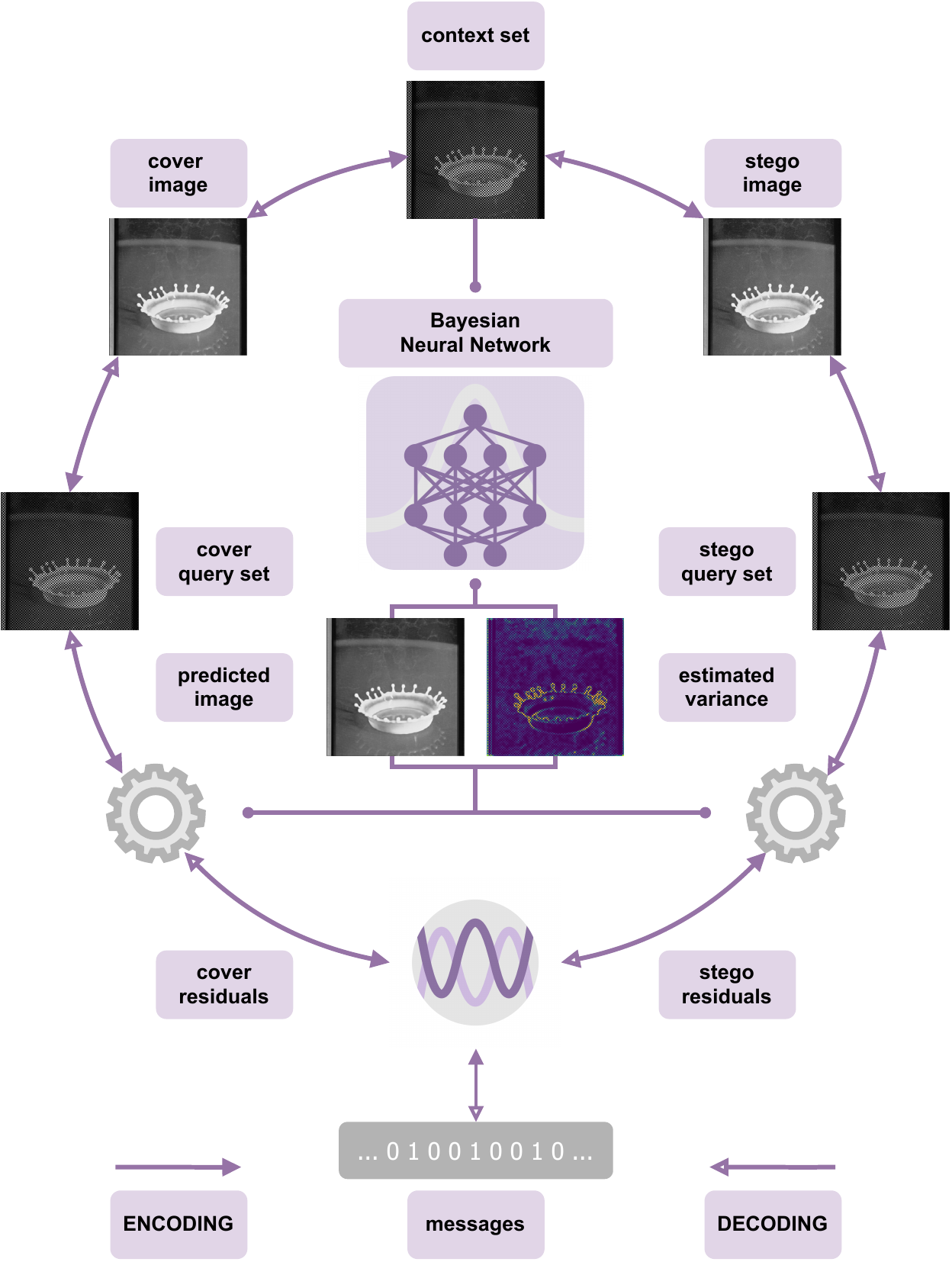}}
\caption{Workflow of reversible steganography with BNN.}
\label{fig:rev_stego}
\end{figure}

Reversible steganography considers the following scenario. A sender communicates a message to a receiver by introducing removable modifications to a carrier signal. We refer to the original signal as the \emph{cover} and its modified counterpart as the \emph{stego}. Residual modulation is a conventional technique used to hide messages within digital images in a reversible fashion. There are many variations of residual modulation~\cite{2005_1381493, 2007_4099409, 2009_4811982, 2011_5762603, 2014_6746082}. An optimal coding for residual modulation is the subject of ongoing research and the choice has few implications for the findings of the study. The following workflow describes a scheme based on residual modulation that incorporates a Bayesian neural network (BNN), as illustrated in Figure~\ref{fig:rev_stego}. In the preliminary phase, pixels of an image are divided into a \emph{context} set and a \emph{query} set, denoted respectively by $\boldsymbol{x}$ and $\boldsymbol{y}$. A common method of context/query splitting is to form a chequerboard pattern such that each query pixel has $4$ adjacent context pixels connected horizontally and vertically. A BNN is then deployed to predict the intensity of each query pixel as well as to estimate the inherent variance in data based on the given contextual information:
\begin{equation}
\{ \hat{\boldsymbol{y}}, \boldsymbol{\sigma}^2 \}= \operatorname{BNN}( \boldsymbol{x} ) .
\end{equation}
For the time being, we assume an uncertainty map as being derived from either/both $\hat{\boldsymbol{y}}$ or/and $\boldsymbol{\sigma}^2$. The map indicates an estimated uncertainty over the prediction for each pixel and provides guidance on message embedding. Uncertainty may be caused by certain rare and stochastic patterns in images. Residual modulation is premised on a statistical principle (i.e. law of error) that residuals generally centre around zero and the frequency of a residual is inversely proportional to its magnitude~\cite{Wilson:1923aa}. In general, it assigns residuals of small magnitude as the \emph{stego channel} to carry the payload at the expense of causing greater distortion to large residuals. Based on the supposition that the expected residual magnitude can be captured by the uncertainty map, we can modulate the residuals in order of ascending uncertainty. In practice, we sort them by ascending uncertainty prior to modulation, leading to an adaptive embedding pathway (in contrast to a sequential or random pathway).
In the encoding phase, residuals are computed by $\boldsymbol{\epsilon} = \boldsymbol{y} - \hat{\boldsymbol{y}}$. 
Message bits are embedded by modulating the residuals, causing recoverable distortion to the residuals:
\begin{equation}
\boldsymbol{\epsilon}^{\prime} = \operatorname{modulate}(\boldsymbol{\epsilon}, \boldsymbol{m}) .
\end{equation}
The modulated residuals are added to the predicted intensities, resulting in a stego query set $\boldsymbol{y}^{\prime} = 
\hat{\boldsymbol{y}} + \boldsymbol{\epsilon}^{\prime}$. 
Message extraction and image restoration are performed in a similar manner. To begin with, the procedures in the preliminary phase are carried out, yielding the same results since the contextual information remains intact. In the decoding phase, the embedding order is identified in accordance with the uncertainty map and the residuals are computed by $\boldsymbol{\epsilon}^{\prime} = \boldsymbol{y}^{\prime} - \hat{\boldsymbol{y}}$. 
The message is extracted and the residuals revert to their pristine state based on the corresponding de-modulation algorithm:
\begin{equation}
\{\boldsymbol{\epsilon}, \boldsymbol{m}\} = \operatorname{demodulate}(\boldsymbol{\epsilon}^{\prime}) .
\end{equation}
Finally, the original image is recovered by $\boldsymbol{y} = \hat{\boldsymbol{y}} + \boldsymbol{\epsilon}$.





\subsection{Bayesian Inference}
A neural network is a non-linear function that maps an input and model parameters to an output:
\begin{equation}
\boldsymbol{y} = f(\boldsymbol{x}, \boldsymbol{\theta}) + \boldsymbol{\epsilon},
\end{equation}
where $\boldsymbol{\epsilon} \sim \mathcal{N}(\boldsymbol{0}, \boldsymbol{\sigma}^2)$ describes Gaussian observation noise. The goal of this regression problem is to find latent model parameters $\boldsymbol{\theta}$ that accurately fit the observed data. Let us denote by $\mathcal{X}$ the training inputs and by $\mathcal{Y}$ the corresponding outputs. Maximum likelihood estimation (MLE) finds the most likely setting of parameters for the set of data, as given by
\begin{equation}
\boldsymbol{\theta}_* = {\underset {\boldsymbol{\theta} }{\operatorname {arg\,max} }}\ \underbrace{p(\mathcal{Y} \mid \mathcal{X}, \boldsymbol{\theta})}_{\textrm{likelihood}} .
\end{equation}
If we have prior knowledge about the distribution of the parameters, we can invoke Bayes' theorem and derive a posterior distribution of parameters: 
\begin{equation}
p(\boldsymbol{\theta} \mid \mathcal{X}, \mathcal{Y})= \frac{ \overbrace{p(\mathcal{Y} \mid \mathcal{X}, \boldsymbol{\theta})}^{\textrm{likelihood}} \cdot \overbrace{p(\boldsymbol{\theta})}^{\textrm{prior}} }{ \underbrace{p(\mathcal{Y} \mid \mathcal{X})}_{\textrm{evidence}} } .
\end{equation}
The denominator of the posterior is the marginal likelihood or model evidence, as defined by
\begin{equation}
p(\mathcal{Y} \mid \mathcal{X}) = \int \underbrace{p(\mathcal{Y} \mid \mathcal{X}, \boldsymbol{\theta})}_{\textrm{likelihood}} \underbrace{p(\boldsymbol{\theta})}_{\textrm{prior}} d \boldsymbol{\theta} .
\end{equation}
Since the denominator does not depend on $\boldsymbol{\theta}$, maximum a posteriori (MAP) estimation ignores it and obtains 
\begin{equation}
\boldsymbol{\theta}_* = {\underset {\boldsymbol{\theta} }{\operatorname {arg\,max} }}\  \underbrace{p(\mathcal{Y} \mid \mathcal{X}, \boldsymbol{\theta})}_{\textrm{likelihood}} \cdot \underbrace{p(\boldsymbol{\theta})}_{\textrm{prior}}.
\end{equation}
From a Bayesian interpretation, finding neural network parameters that minimise a loss function is conceptually similar to MLE, whereas training with weight-decay regularisation has a similar underlying principle to MAP~\cite{10.5555/1162264, 10.5555/2380985, Goodfellow-et-al-2016}.
Bayesian inference takes the full posterior distribution into account to support robust decision-making. Rather than relying solely on a single hypothesis (i.e. a specific setting of parameters), Bayesian inference leverages all possible settings of parameters, weighted by their plausibilities (i.e. posterior probabilities)~\cite{deisenroth2020mathematics}. It propagates uncertainty from the parameters to the data by deriving the (posterior) \emph{predictive distribution} of $\boldsymbol{y}_*$ at a test input $\boldsymbol{x}_*$ using the parameter posterior:
\begin{equation}
p(\boldsymbol{y}_* \mid \boldsymbol{x}_*, \mathcal{X}, \mathcal{Y}) = \int \underbrace{p(\boldsymbol{y}_* \mid \boldsymbol{x}_*, \boldsymbol{\theta})}_{\textrm{likelihood}} \underbrace{p(\boldsymbol{\theta} \mid \mathcal{X}, \mathcal{Y})}_{\textrm{posterior}} d \boldsymbol{\theta} .
\end{equation}
The parameter posterior involves the computation of model evidence, which requires solving an integration referred to as \emph{marginalisation}. However, marginalisation is analytically intractable for deep-learning models and therefore we have to resort to approximation techniques.


\subsection{Monte Carlo Dropout}
\begin{figure}[t]
\centerline{\includegraphics[width=0.95\columnwidth]{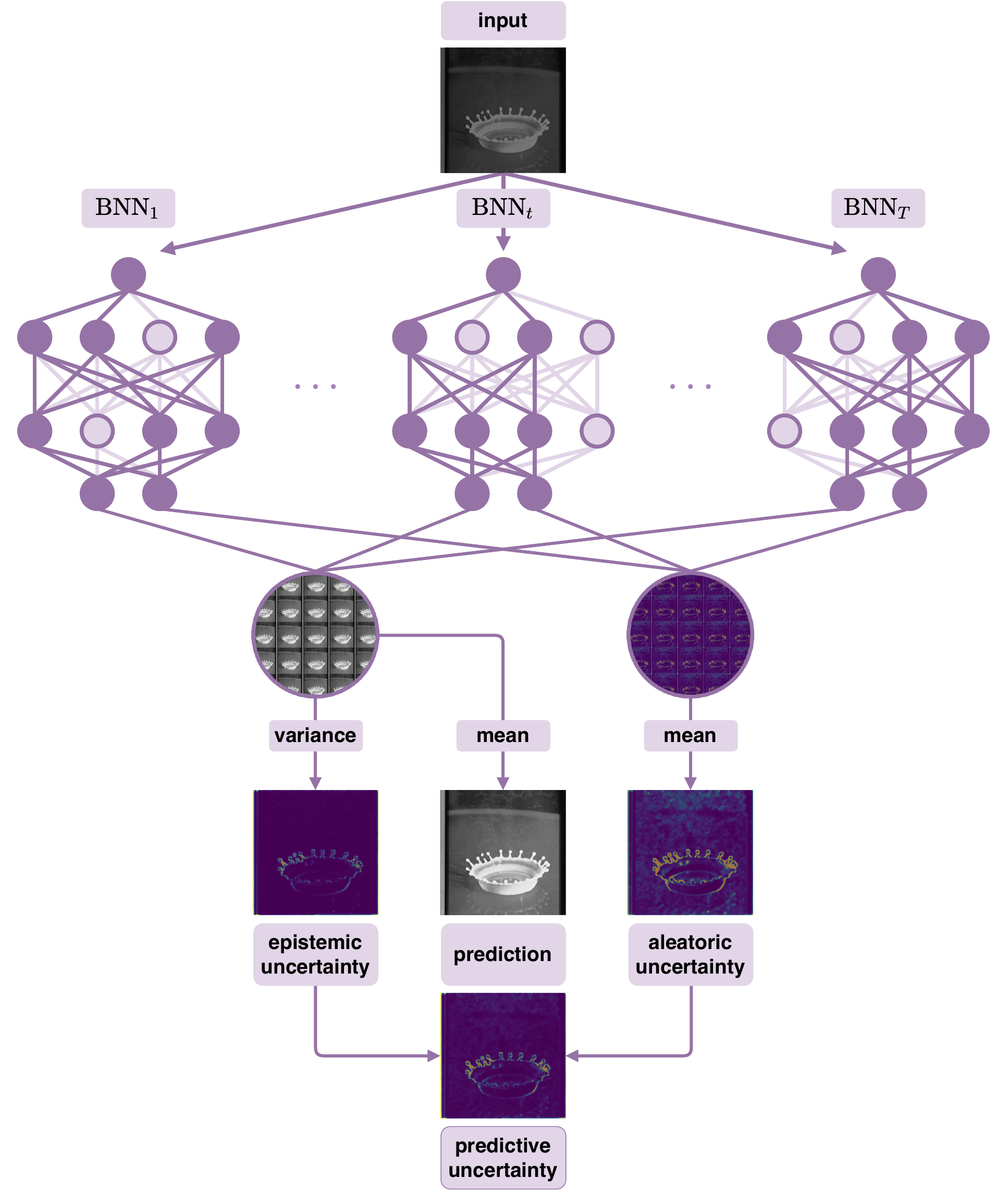}}
\caption{Monte Carlo dropout as stochastic forward passes through BNN.}
\label{fig:BNN}
\end{figure}

Variational inference is a technique for approximating intractable integrals in Bayesian inference~\cite{10.1145/168304.168306, Jordan:1999aa, DBLP:journals/corr/KingmaW13, 10.5555/2986459.2986721, 10.5555/3045118.3045290}. Instead of evaluating the parameter posterior, we approximate it with a variational distribution $q(\boldsymbol{\theta})$, which belongs to a family of distributions of simpler form. By replacing $p(\boldsymbol{\theta} \mid \mathcal{X}, \mathcal{Y})$ with $q(\boldsymbol{\theta})$ in the predictive distribution and approximating the integral with Monte Carlo integration, we derive that
\begin{equation} 
\begin{split}
p(\boldsymbol{y}_* \mid \boldsymbol{x}_*, \mathcal{X}, \mathcal{Y}) &\stackrel{\textrm{VI}}{\approx} \int p(\boldsymbol{y}_* \mid \boldsymbol{x}_*, \boldsymbol{\theta}) q(\boldsymbol{\theta}) d \boldsymbol{\theta} \\ 
&\stackrel{\textrm{MC}}{\approx} \frac{1}{T} \sum_{t=1}^T p(\boldsymbol{y}_* \mid \boldsymbol{x}_*, \hat{\boldsymbol{\theta}}_t) ,
\end{split}
\end{equation}
where $\hat{\boldsymbol{\theta}}_t \sim q(\boldsymbol{\theta})$. Sampling model parameters from a variational distribution can be simulated by dropout~\cite{gal_iclr2016}, a stochastic process of multiplying the output of each neurone by a random variable drawn from a Bernoulli distribution~\cite{2014_Dropout}. In other words, each dropout configuration deactivates a portion of neurones, yielding a plausible realisation of the parametric model. Performing stochastic forward passes $T$ times through a model with different dropout masks results in an ensemble of neural networks, each with a slightly different sparse graph structure. The usage of dropout at the inference stage is referred to as \emph{Monte Carlo dropout}~\cite{pmlr-v48-gal16}, as illustrated in Figure~\ref{fig:BNN}. 

\subsection{Uncertainty Disentanglement}
Bayesian inference offers a predictive distribution from which we can derive \emph{predictive uncertainty}. This is the sum of uncertainty due to observation noise and uncertainty in the model parameters. According to the law of total variance, a variance can be decomposed into `unexplained' and `explained' components. This permits the disentanglement of \emph{aleatoric} and \emph{epistemic} uncertainties from predictive uncertainty~\cite{10.5555/3295222.3295309}:
\begin{equation}
\mathbb{V}(\boldsymbol{y} \mid \boldsymbol{x}) = \underbrace{\mathbb{E}[\mathbb{V}(\boldsymbol{y} \mid \boldsymbol{x}, \boldsymbol{\theta})]}_{\textrm{aleatoric}} + \underbrace{\mathbb{V}(\mathbb{E}[\boldsymbol{y} \mid \boldsymbol{x}, \boldsymbol{\theta}])}_{\textrm{epistemic}} .
\end{equation}
Aleatoric uncertainty captures randomness (or noise) inherent in observations:
\begin{equation}
\mathbb{E}[\boldsymbol{\sigma}^2] \approx \frac{1}{T} \sum_{t=1}^T \boldsymbol{\sigma}^2_t .
\end{equation}
Epistemic uncertainty occurs due to limited knowledge (or training data):
\begin{equation}
\begin{split}
\mathbb{V}(\hat{\boldsymbol{y}}) &= \mathbb{E}[(\hat{\boldsymbol{y}} - \mathbb{E}[\hat{\boldsymbol{y}}])^2]
= \mathbb{E}[\hat{\boldsymbol{y}}^2] - \mathbb{E}[\hat{\boldsymbol{y}}]^2 \\
&\approx \frac{1}{T} \sum_{t=1}^T \hat{\boldsymbol{y}}_t^2 - (\frac{1}{T}\sum_{t=1}^T \hat{\boldsymbol{y}}_t)^2 .
\end{split}
\end{equation}
Hybrid predictive uncertainty can be computed by adding together aleatoric and epistemic uncertainties, each normalised with its sum. We perform dropout at test time and sample i.i.d. results $\{\hat{\boldsymbol{y}}_t, \boldsymbol{\sigma}^2_t\}_{t=1}^T \sim \operatorname{BNN}_{\hat{\boldsymbol{\theta}}_t}(\boldsymbol{x})$. The loss function for this \emph{dual-headed} BNN is composed of a distance function $\mathcal{D}$ and a regulariser $\mathcal{R}$ balanced by $\lambda$~\cite{374138}, as given by
\begin{equation}
\mathcal{L}_{\operatorname{BNN}} = \mathcal{D}(\boldsymbol{y}, \hat{\boldsymbol{y}}, \boldsymbol{\sigma}^2) + \lambda\mathcal{R}(\boldsymbol{\sigma}^2) .
\end{equation}
Let us denote by $N$ the number of pixels in an image. The distance function is defined by
\begin{equation}
\mathcal{D} = \frac{1}{N} \sum_{n=1}^N (y_{n} - \hat{y}_{n} )^2 \left(\frac{\sigma_{n}^2}{\operatorname{sum}(\boldsymbol{\sigma}^2)}\right)^{-1} ,
\end{equation}
where the first term is the Euclidean distance and the second term represents the inverse of the normalised variance. This weighted distance term discourages the model from causing high regression residuals with low uncertainty and attenuates loss in conditions of high uncertainty.
The regulariser is defined by
\begin{equation}
\mathcal{R} = \frac{1}{N} \sum_{n=1}^N \ln\sigma_{n}^2 .
\end{equation}
This regularisation term is designed to penalise a high degree of uncertainty, thereby preventing the model from inactive learning.
\begin{figure}[t]
\centerline{\includegraphics[width=0.95\columnwidth]{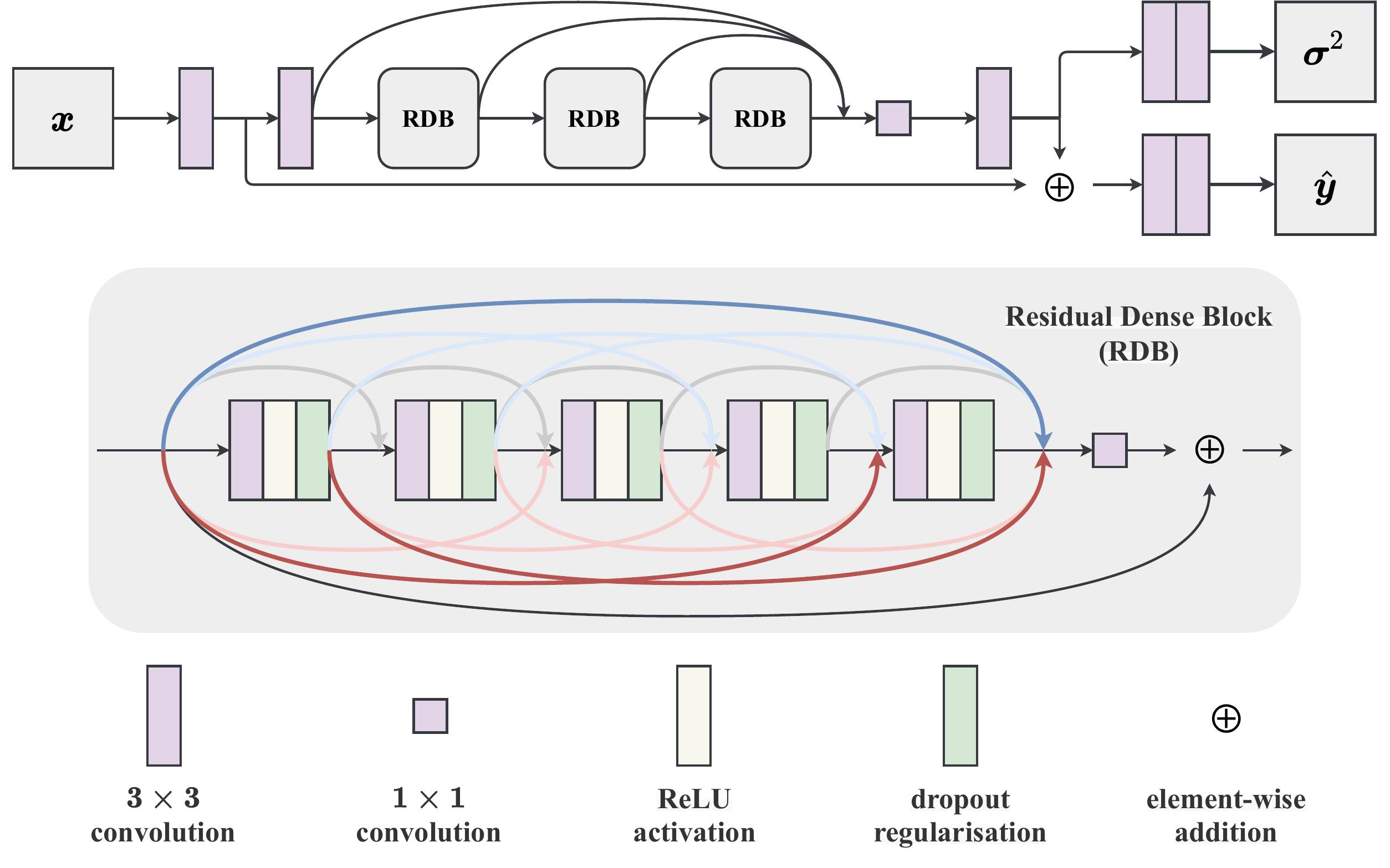}}
\caption{Architecture of dual-headed RDN with dropout layers.}
\label{fig:RDN_dual}
\end{figure}

\section{Experiments}
The primary purpose of our experiments is to identify the contribution made by Bayesian uncertainty analysis to steganographic performance. We employ an advanced predictive neural network as a baseline model for benchmarking and build an uncertainty-aware model therefrom.

\subsection{Experimental Setup}
Our implementation of the BNN is based primarily on the residual dense network (RDN), which is a state-of-the-art model for low-level computer vision tasks (e.g. image reconstruction, super-resolution, and denoising)~\cite{2018_8578360}. We use this model as a baseline predictive model and build an uncertainty-aware predictive model upon it. The RDN model is characterised by a tangled labyrinth of residual and dense connections. We train the RDN model on the BOSSbase dataset~\cite{2011_BOSSbase}, which consists of 10,000 greyscale photographs collected for an academic competition in the field of digital steganography. The inference set comprises standard test images from the USC-SIPI dataset~\cite{2006_USC_SIPI}. For the uncertainty-aware model, we apply a dropout layer after each non-linear activation function and stack two output branches to form a dual-headed model, as illustrated in Figure~\ref{fig:RDN_dual}. The dropout rate is set to 0.3, the loss balancing parameter $\lambda$ to 1, and the number of dropout samples $T$ to 1,000 empirically. 

%
%

\begin{figure*}[t!] 
\centering
\subfloat[input]{\includegraphics[width=0.28\columnwidth]{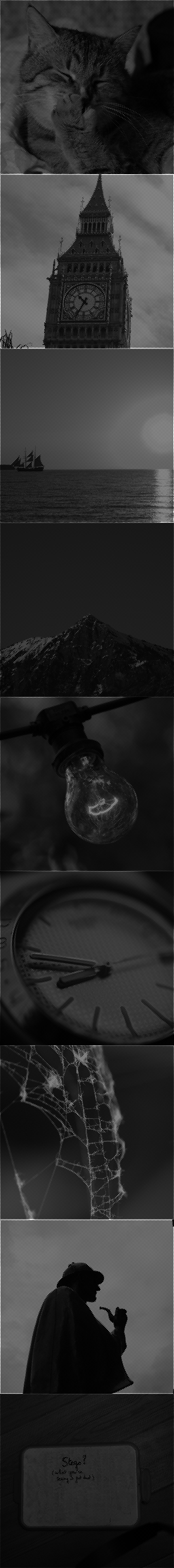}}
\hfil
\subfloat[target]{\includegraphics[width=0.28\columnwidth]{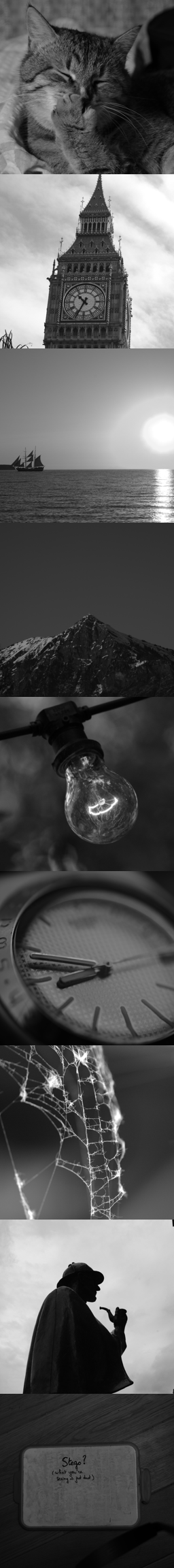}}
\hfil
\subfloat[output]{\includegraphics[width=0.28\columnwidth]{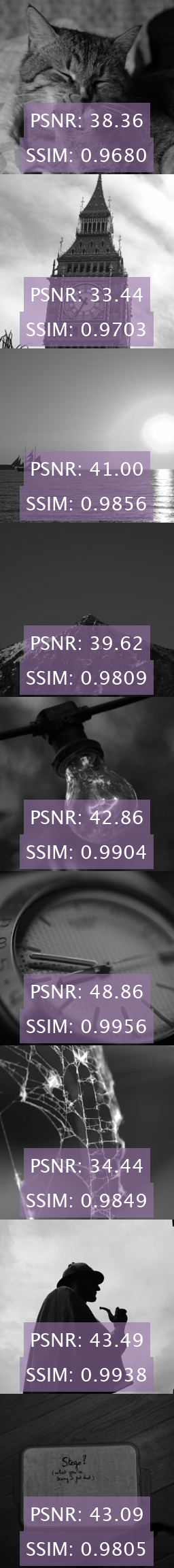}}
\hfil
\subfloat[aleatoric]{\includegraphics[width=0.28\columnwidth]{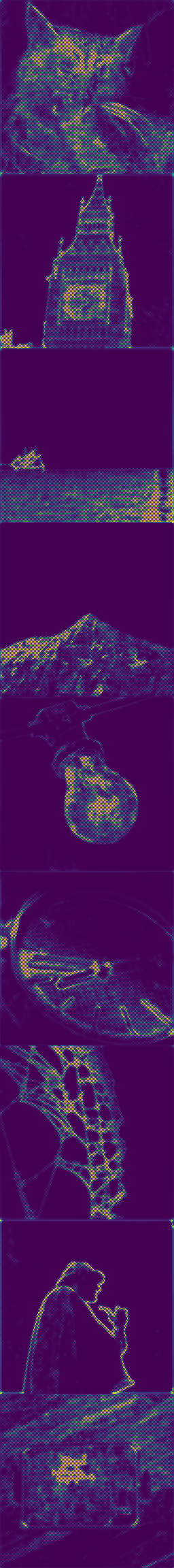}}
\hfil
\subfloat[epistemic]{\includegraphics[width=0.28\columnwidth]{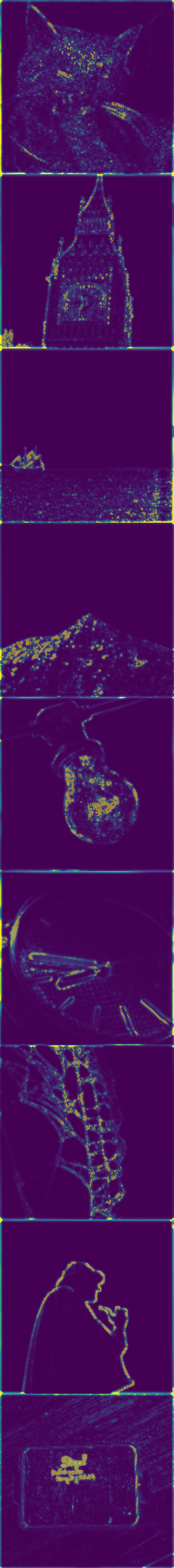}}
\hfil
\subfloat[predictive]{\includegraphics[width=0.28\columnwidth]{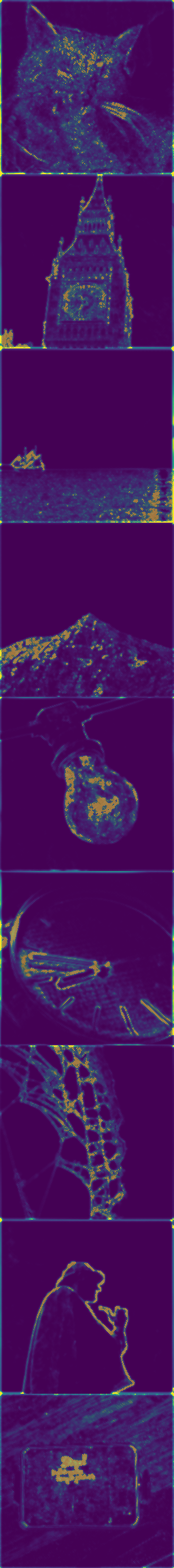}}

\caption{Visual results from BOSSbase dataset.}
\label{fig:montage_BOSS}
\end{figure*}

%
%

\begin{figure*}[t!] 
\centering
\subfloat[input]{\includegraphics[width=0.28\columnwidth]{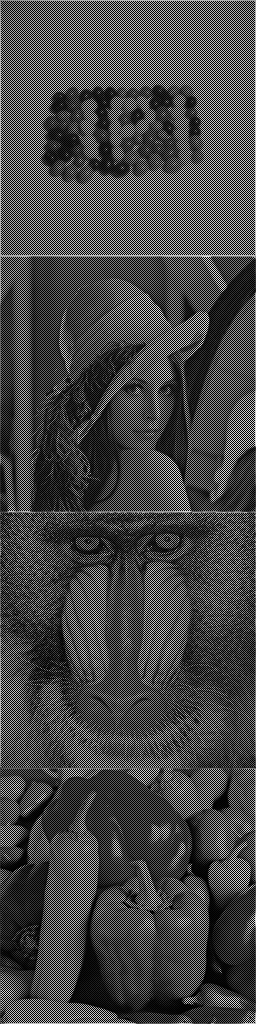}}
\hfil
\subfloat[target]{\includegraphics[width=0.28\columnwidth]{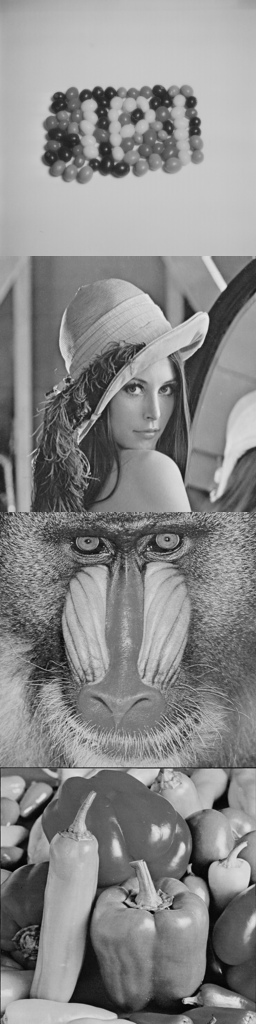}}
\hfil
\subfloat[output]{\includegraphics[width=0.28\columnwidth]{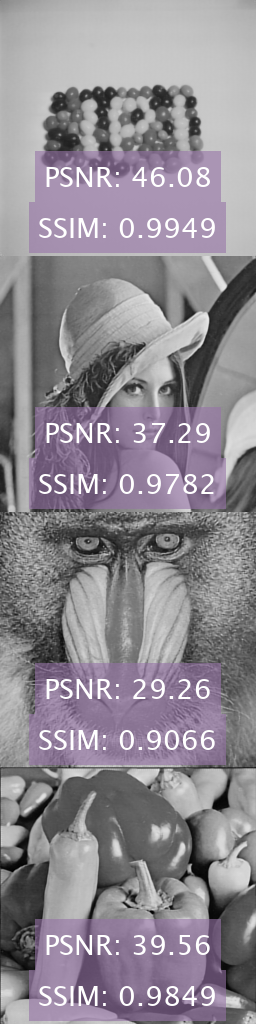}}
\hfil
\subfloat[aleatoric]{\includegraphics[width=0.28\columnwidth]{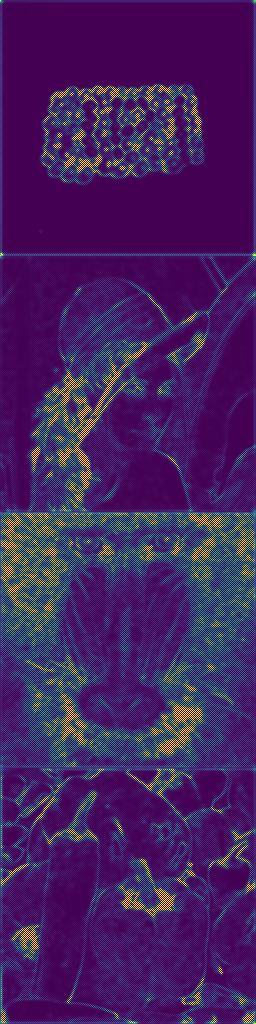}}
\hfil
\subfloat[epistemic]{\includegraphics[width=0.28\columnwidth]{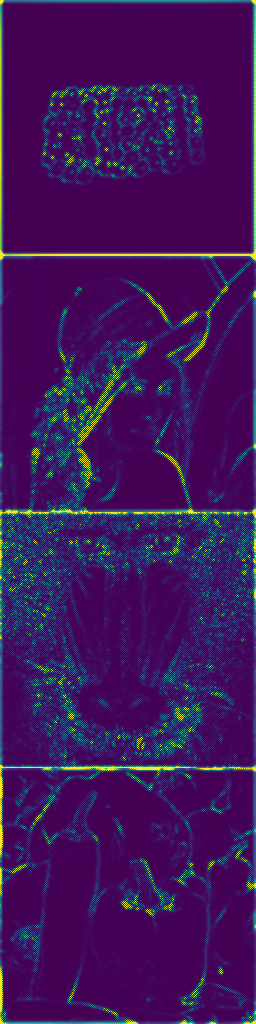}}
\hfil
\subfloat[predictive]{\includegraphics[width=0.28\columnwidth]{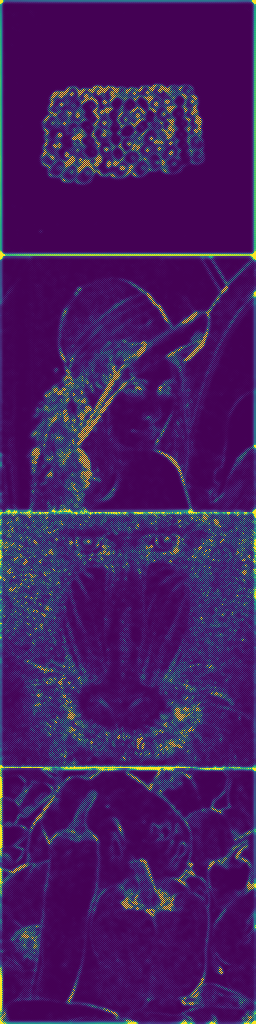}}

\caption{Visual results from USC-SIPI dataset.}
\label{fig:montage_USCSIPI}
\end{figure*}


%
%
%

\begin{figure*}[t!] 
\centering
\subfloat[Jelly Beans]{\includegraphics[width=0.45\columnwidth]{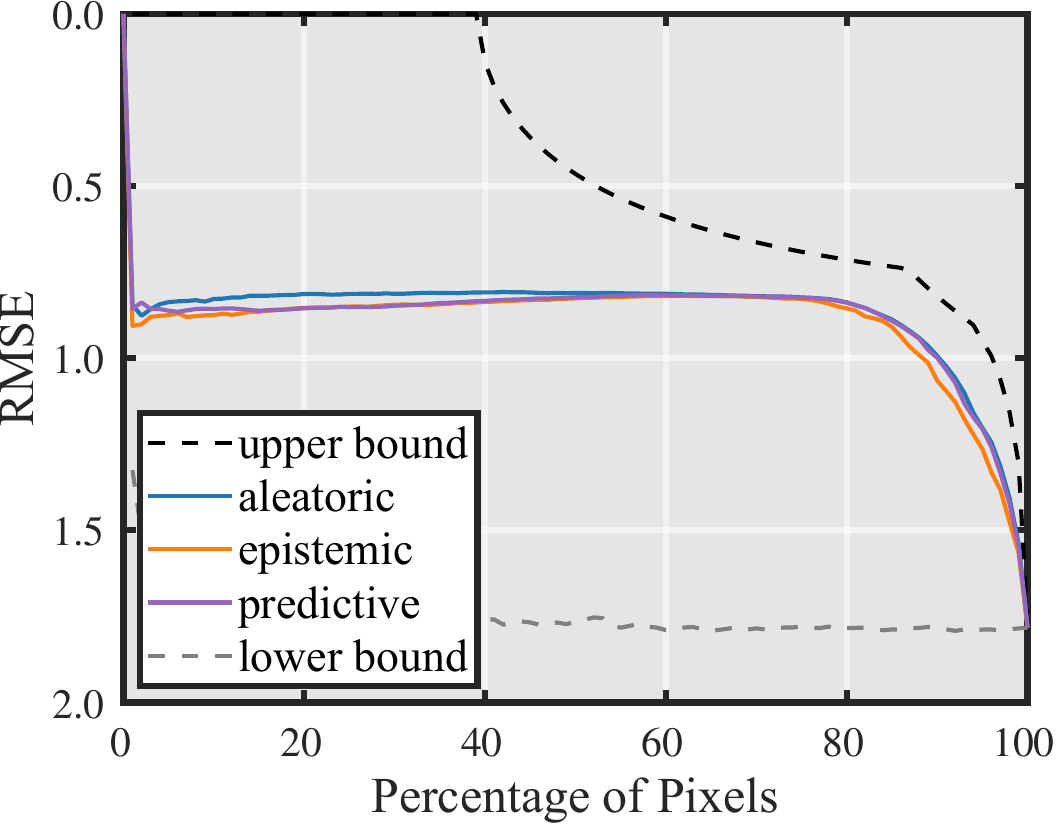}}
\hfil
\subfloat[Lena]{\includegraphics[width=0.45\columnwidth]{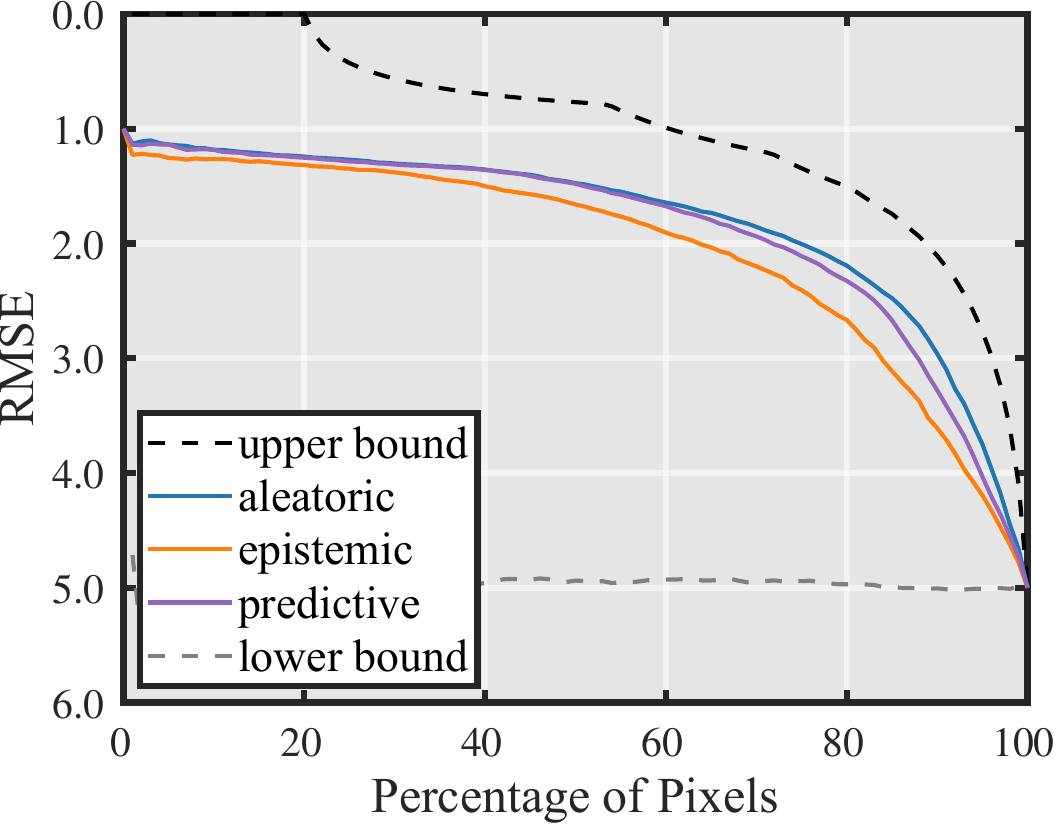}}
\hfil
\subfloat[Mandrill]{\includegraphics[width=0.45\columnwidth]{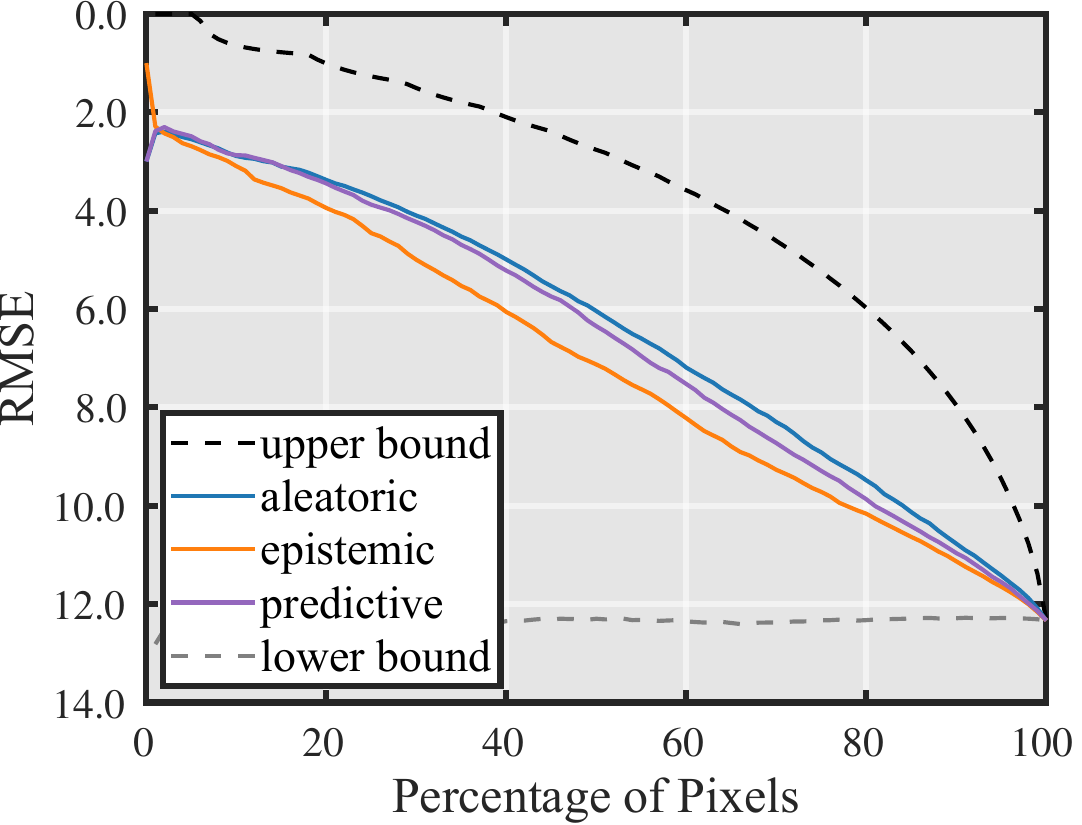}}
\hfil
\subfloat[Peppers]{\includegraphics[width=0.45\columnwidth]{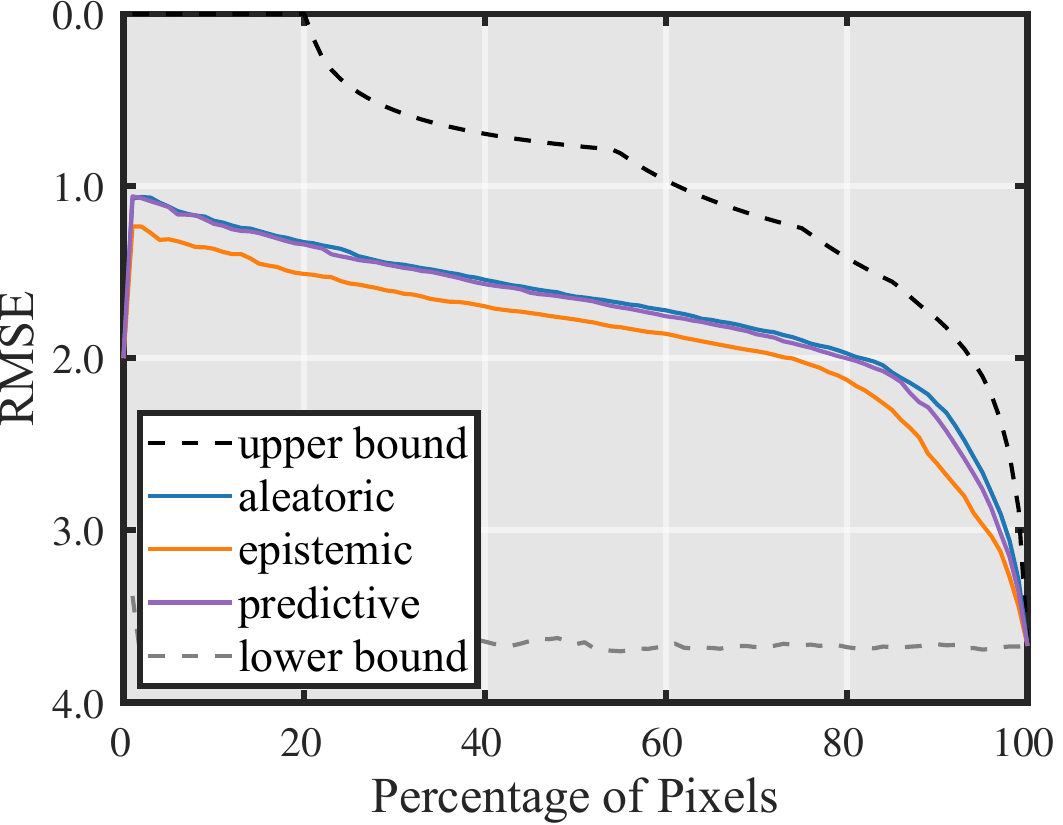}}

\caption{Uncertainty quantification performance by RMSE vs percentage of pixels.}
\label{fig:rmse}
\end{figure*}

\begin{figure*}[t!] 
\centering
\subfloat[Jelly Beans]{\includegraphics[width=0.45\columnwidth]{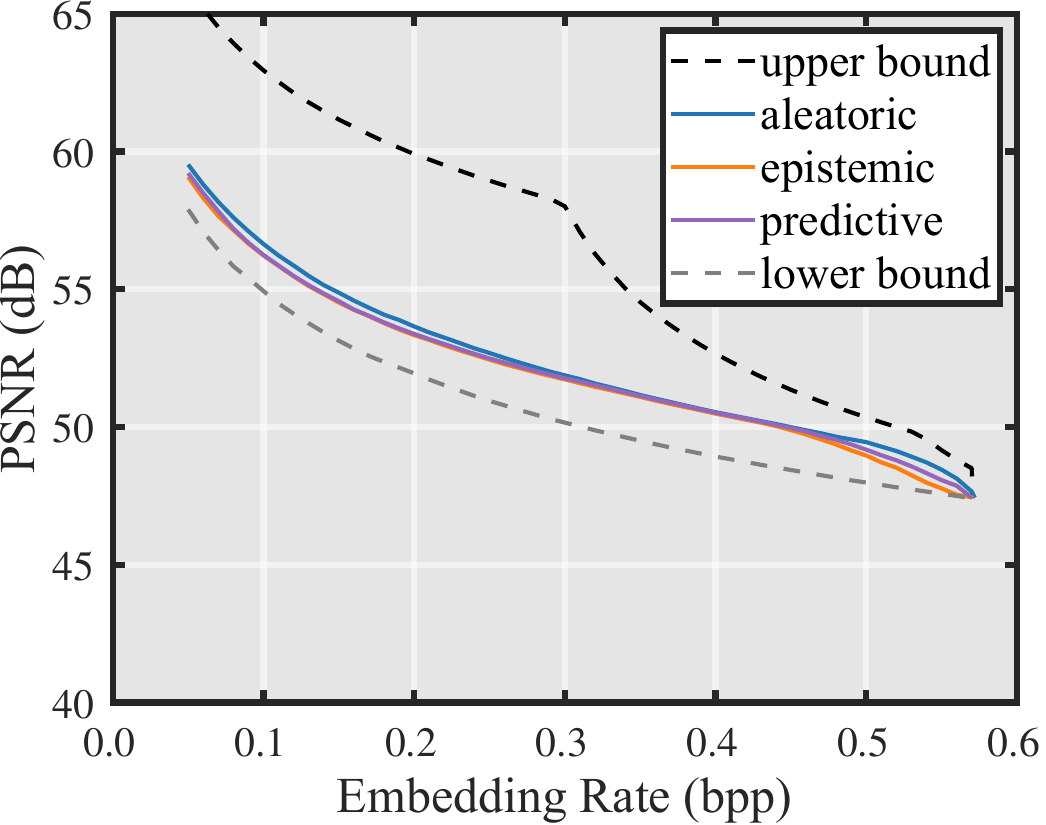}}
\hfil
\subfloat[Lena]{\includegraphics[width=0.45\columnwidth]{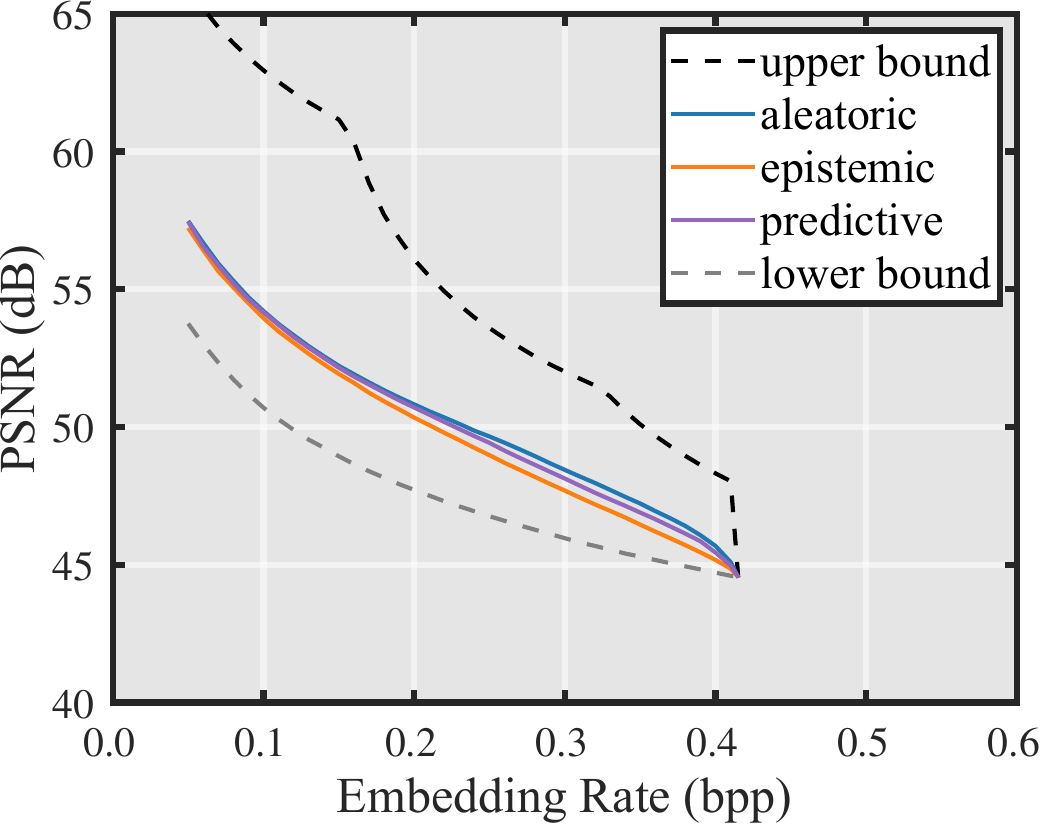}}
\hfil
\subfloat[Mandrill]{\includegraphics[width=0.45\columnwidth]{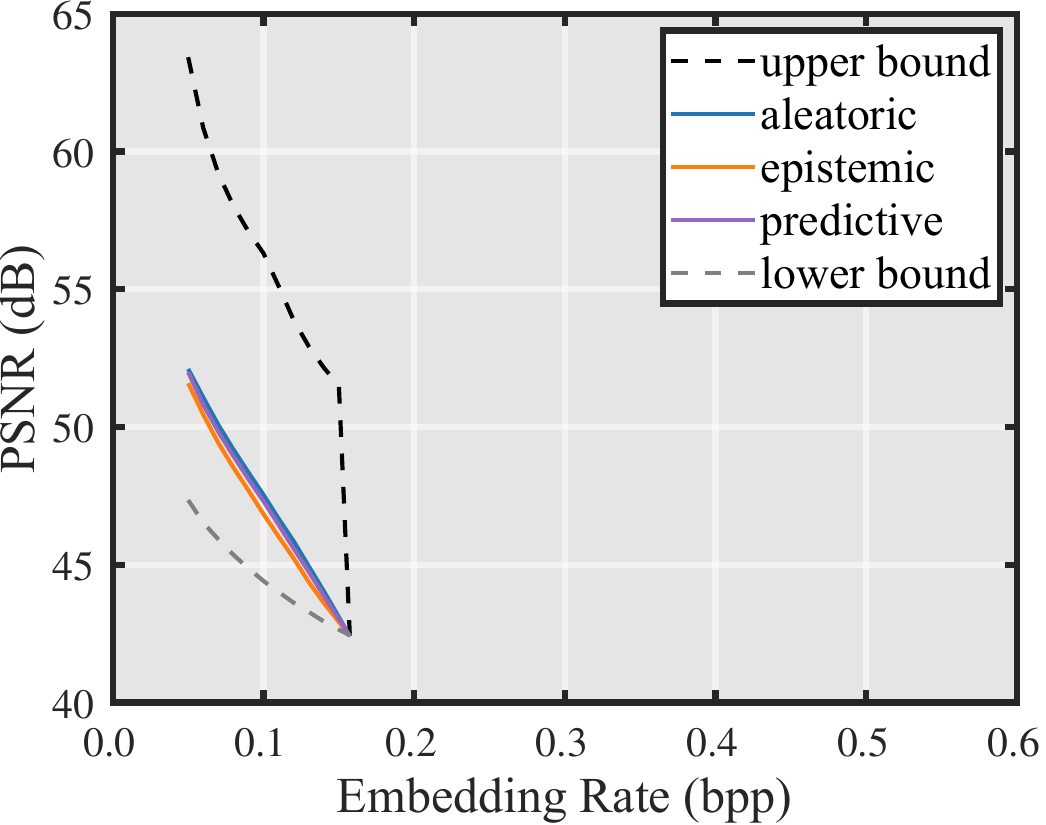}}
\hfil
\subfloat[Peppers]{\includegraphics[width=0.45\columnwidth]{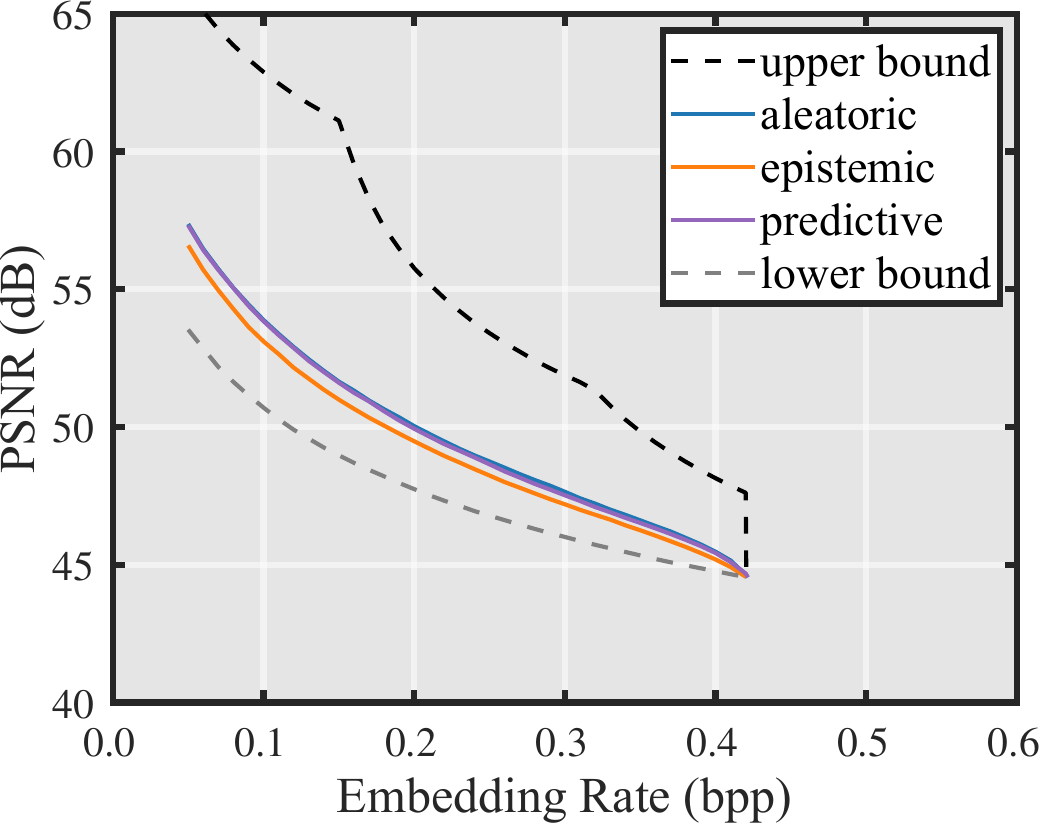}}

\caption{Steganographic performance by PSNR vs embedding rate.}
\label{fig:rate_dist}
\end{figure*}

%

\subsection{Experimental Results}
Figures~\ref{fig:montage_BOSS} and~\ref{fig:montage_USCSIPI} visualise the experimental results. The metrics used for measuring the quality of predicted images are peak signal-to-noise ratio (PSNR) and structural similarity (SSIM). In general, predictive performance is high for smooth images and low for richly textured images. The chequerboard artefacts are observed in uncertainty maps since a chequered pattern (derived from $4$-connectivity) is used for context-query splitting. The intensities of context pixels in an input image are consistent with those in a corresponding target image. The results suggest that the model is capable of eliminating uncertainty regarding the context pixels. It can be seen that uncertainty is concentrated around edges, contours, rare patterns and textural details. Figure~\ref{fig:rmse} depicts uncertainty quantification performance, represented by the root-mean-square error (RMSE) vis-\`{a}-vis the percentage of pixels. It illustrates the deviation of predictions as the percentage of pixels increases, where the pixels are selected in ascending order of uncertainty magnitude. The upper bound is obtained by selecting pixels in ascending order of residual magnitude, whereas the lower bound is computed by selecting pixels in random order. The results verify the validity of uncertainty analysis by comparing the uncertainty-aware selection with the random selection. A more accurate uncertainty measurement is expected to produce a curve closer to the upper bound. Figure~\ref{fig:rate_dist} evaluates steganographic performance with rate-distortion curves. Capacity is measured by the embedding rate expressed in bits per pixel (bpp) and distortion is measured by the PSNR expressed in decibels (dB). The results show that an adaptive embedding pathway derived by Bayesian uncertainty analysis leads to a better rate-distortion performance than a random embedding pathway.

\section{Conclusion}
\label{sec:con}
In this paper, we study reversible steganography with deep learning and analyse uncertainty in predictive models based upon a Bayesian framework. We apply the Monte Carlo dropout to approximate the predictive distribution and derive aleatoric and epistemic uncertainties therefrom. A dual-headed neural network is constructed for estimating uncertainty in an unsupervised manner. Experimental results demonstrate state-of-the-art steganographic performance benchmarked against a non-Bayesian baseline, confirming the contribution of uncertainty analysis. We hope that this article can contribute to future research on reversible steganography and envisage further progress being ushered in with new developments of Bayesian deep learning.


\bibliographystyle{Transactions-Bibliography/IEEEtran}
\bibliography{./Bib/myBib_abbrv}

\begin{IEEEbiography}[{\includegraphics[width=1in,height=1.25in,clip,keepaspectratio]{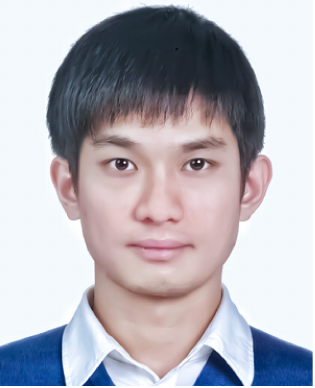}}]{Ching-Chun Chang}
received his PhD in Computer Science from the University of Warwick, UK, in 2019. He engaged in a short-term scientific mission supported by European Cooperation in Science and Technology Actions at the Faculty of Computer Science, Otto von Guericke University Magdeburg, Germany, in 2016. He was granted the Marie-Curie fellowship and participated in a research and innovation staff exchange scheme supported by Marie Sk\l{}odowska-Curie Actions at the Department of Electrical and Computer Engineering, New Jersey Institute of Technology, USA, in 2017. He was a Visiting Scholar at the School of Computer and Mathematics, Charles Sturt University, Australia, in 2018, and at the School of Information Technology, Deakin University, Australia, in 2019. He was a Research Fellow at the Department of Electronic Engineering, Tsinghua University, China, in 2020. He has been a Postdoctoral Fellow at the National Institute of Informatics, Japan, since 2021. His research interests include steganography, watermarking, forensics, biometrics, cybersecurity, applied cryptography, image processing, computer vision, natural language processing, computational linguistics, machine learning and artificial intelligence.
\end{IEEEbiography}

\EOD

\end{document}